# A Tendon-Driven Five-Fingered Hand with Distributed Tactile Perception for Dexterous Manipulation

Huayang Chen, Longhui Qin*

School of Mechanical Engineering, Southeast University, Nanjing, China
*Corresponding author: Longhui Qin (lhqin@seu.edu.cn)

**Abstract**. To apply the techniques of embodied artificial intelligence to humanoid robots for complex manipulations, dexterous robotic hands are indispensable, which are restricted by the dexterity and tactile perception capability. In this work, we proposed a novel design of tendon-driven five-fingered hand with distributed tactile perception. With a soft-rigid-hybrid structure employed, both compliance and operational force are endowed to the hand. Dual-modality tactile sensing elements are distributed on the distal and middle phalanges of all five fingers, enabling the simultaneous detection of static contact and dynamic force variations. Manipulation experiments, including counting gestures, finger-to-thumb pinching, object grasping, and bottle-grasp tactile recording, demonstrate the feasibility of the integrated actuation-perception system.

## 1. Introduction

Relying on our dexterous hands, human is competent to various complex manipulations and endowed with the capability of rapid adaptation in strange and unstructured environments. Although the single Degree-of-Freedom (DOF) grippers have been widely adopted in the community of robotics [1], [2], it is restricted to simple operations of regular-shape objects in low-complexity tasks. Thus, in occasions that require in-hand manipulation or delicate operations, it is nontrivial to design dexterous hands for tool use and fine adjustment of object pose [3], [4].

In the design of robotic hands, numerous issues should be taken into consideration, such as employed material, mechanical structure, driving mode, perception and son on. As per the employed material, existing robotic hands can be primarily divided into three categories: rigid, soft and their hybrid structure. A representative of rigid hands is the Shadow hand, and it can be applied to object reorientation [5]. The opens-source project DexHand also showed high dexterity [6]. However, rigid hands usually suffer from high control complexity and low compliance in grasping flexible objects. Soft hands possessed much higher flexibility in practical applications [7], while it failed to provide sufficient manipulation force. How to combine the advantages of both rigid and soft hands remains challenging. Meanwhile, tactile sensing is absent in most existing hands, which further limits their applications.

To address these challenges, this paper presents an integrated tendon-driven five-fingered robotic hand with distributed dual-modality tactile perception. Compared with conventional tactile grippers that mainly arrange sensing elements at limited contact regions, the proposed system distributes piezoresistive and piezoelectric sensing elements on both the distal and middle phalanges of all five fingers. The main contributions are threefold. First, a soft-rigid-hybrid five-fingered hand is developed with tendon-driven finger flexion, finger swing, and wrist motion. Second, strain-gauge and PVDF sensing elements are integrated into multiple phalanges to perceive static contact and dynamic force variations. Third, an integrated actuation-perception architecture is established to synchronously execute motion commands and record multi-channel tactile signals during manipulation.

## 2. Hand Design and System Integration

### 2.1 Overall Architecture of the Dexterous Hand

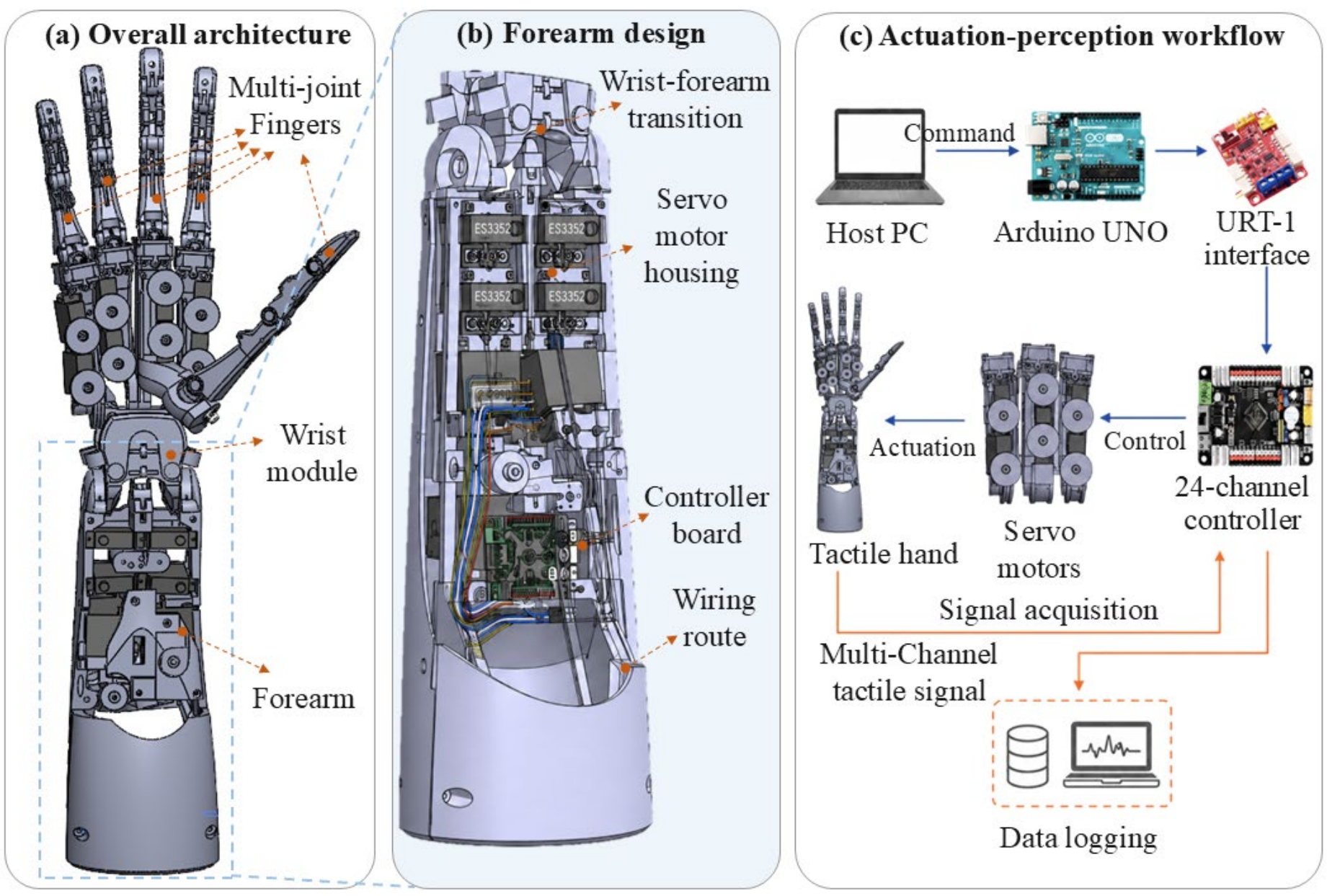

Fig. 1. System architecture and actuation-perception workflow of the tendon-driven five-fingered hand. (a) CAD (computer aided design) model of the entire hand system consisting of five fingers, a palm, a wrist module and a forearm. (b) Detailed view of the forearm. (c) Actuation and perception workflow of the dexterous hand system.

Inspired by the open-source project DexHand [6], we adopted a similar architecture to construct the hand structure. It was designed as tendon-driven mode considering its compliance in manipulations and underactuated coupling is considered between adjacent fingers. As summarized in Fig. 1(a), the entire hand system consisted of five fingers, a palm, a wrist module and a forearm. The multi-joint fingers were arranged mimicking the layout of human fingers. A wrist-forearm transition module, servo motor housing and a controller board were integrated into the forearm, as shown in Fig. 1(b), in order to reducing the mass and simplifying the cumbersome structure of distal hand part while preserving the access to tendon pretensioning, wiring inspection, and hardware adjustment. In addition, the cable routing problem need be taken into full consideration as both actuation and perception cables were contained to transmit respective signals. It should be noted that our proposed hand design more resemble human hand in the structure and functionality since a soft-rigid-hybrid structure was employed and distributed tactile sensory system was configured. It has been revealed that soft-rigid-hybrid structured fingers could provide more compliance in delicate operations [8].

## 2.2 Actuation and Embedded Command Flow

The control and acquisition architecture comprised forearm-mounted actuator units, embedded control hardware, interface electronics, and a multi-channel tactile acquisition unit into a unified actuation-perception pipeline, as illustrated in Fig. 1(c). Host-side commands are parsed by the embedded control unit, assigned to actuator targets, and transmitted through the interface electronics. During execution, command states and tactile frames are acquired in parallel so that mechanical motion can be analyzed together with the corresponding contact response.

A state-based motion-primitive strategy is adopted to organize tendon-driven manipulation. Low-level commands specify the actuator target and transition time, and are grouped into grasp, hold, and release phases. It supports repeatable examination of the relationships among tendon-driven motions, contact formation, and tactile response.

## 2.3 Finger Mechanism and Tendon-Driven Transmission

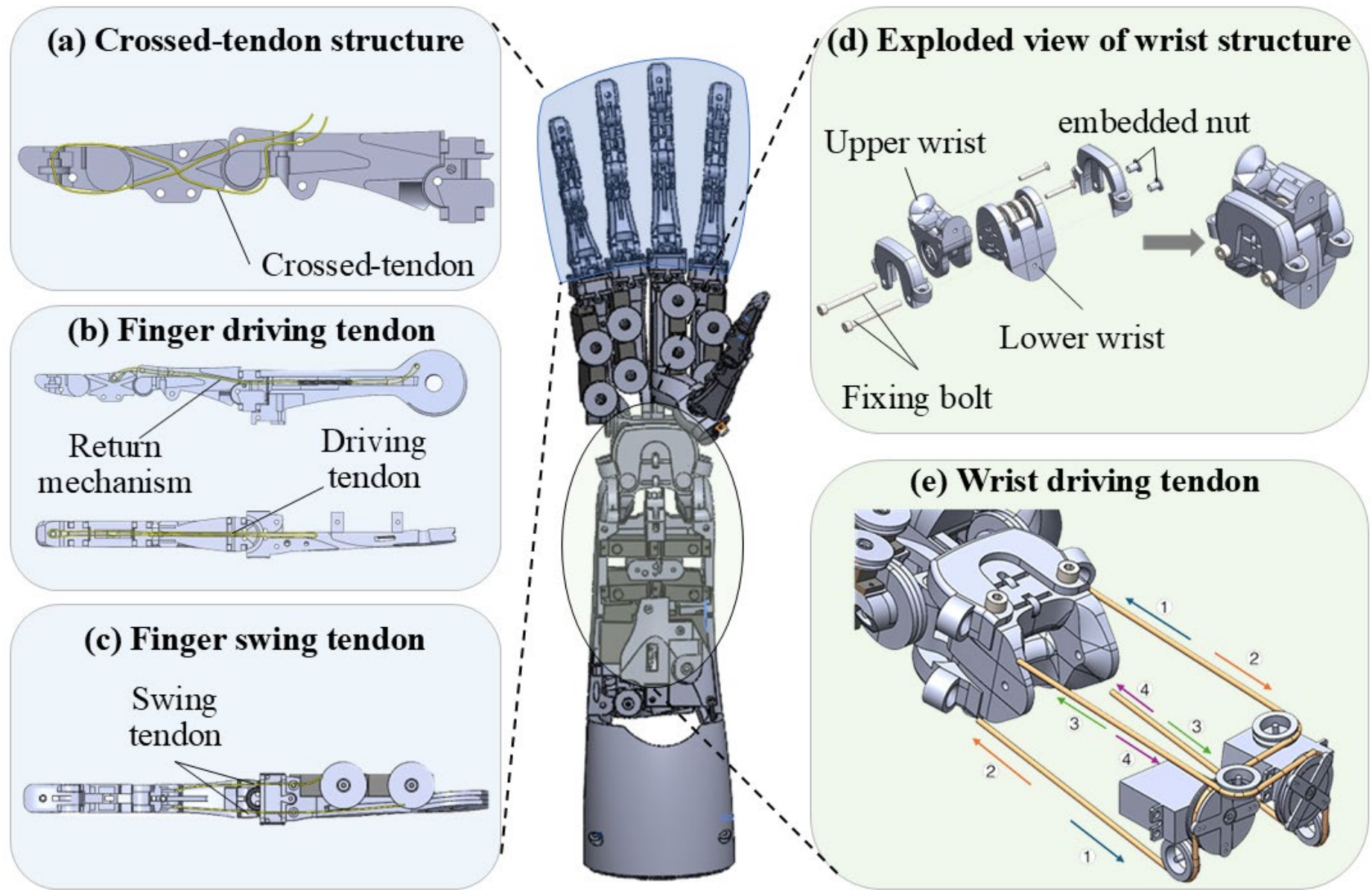

Fig. 2. Transmission module design of the dexterous hand.

Each finger is treated as a coupled joint chain whose actual posture is jointly determined by tendon displacement, routing geometry, pretension, and contact conditions rather than by an independently controlled joint trajectory. The thumb was arranged to support opposing motions, while the other fingers are configured for coordinated closure during grasping and pinch-like operations.

As shown in Fig. 2, crossed-tendon paths, finger driving/swing tendons, and wrist-driving tendons constitute the transmission system of the dexterous hand. In case of free-space closure, the fingers follow nominal coupled trajectories. Once contact occurs, the relevant tendons would be driven to actuate the corresponding joints move towards less-constrained directions, therefore producing adaptive wrapping behaviors. The hand status differs for different contact scenarios. Besides, the routing geometry and pretension directly influence the repeatability and contact formation.

## 3. Tactile Sensing and Contact Interpretation

### 3.1 Design of the Tactile Sensing Module in Dexterous Hand

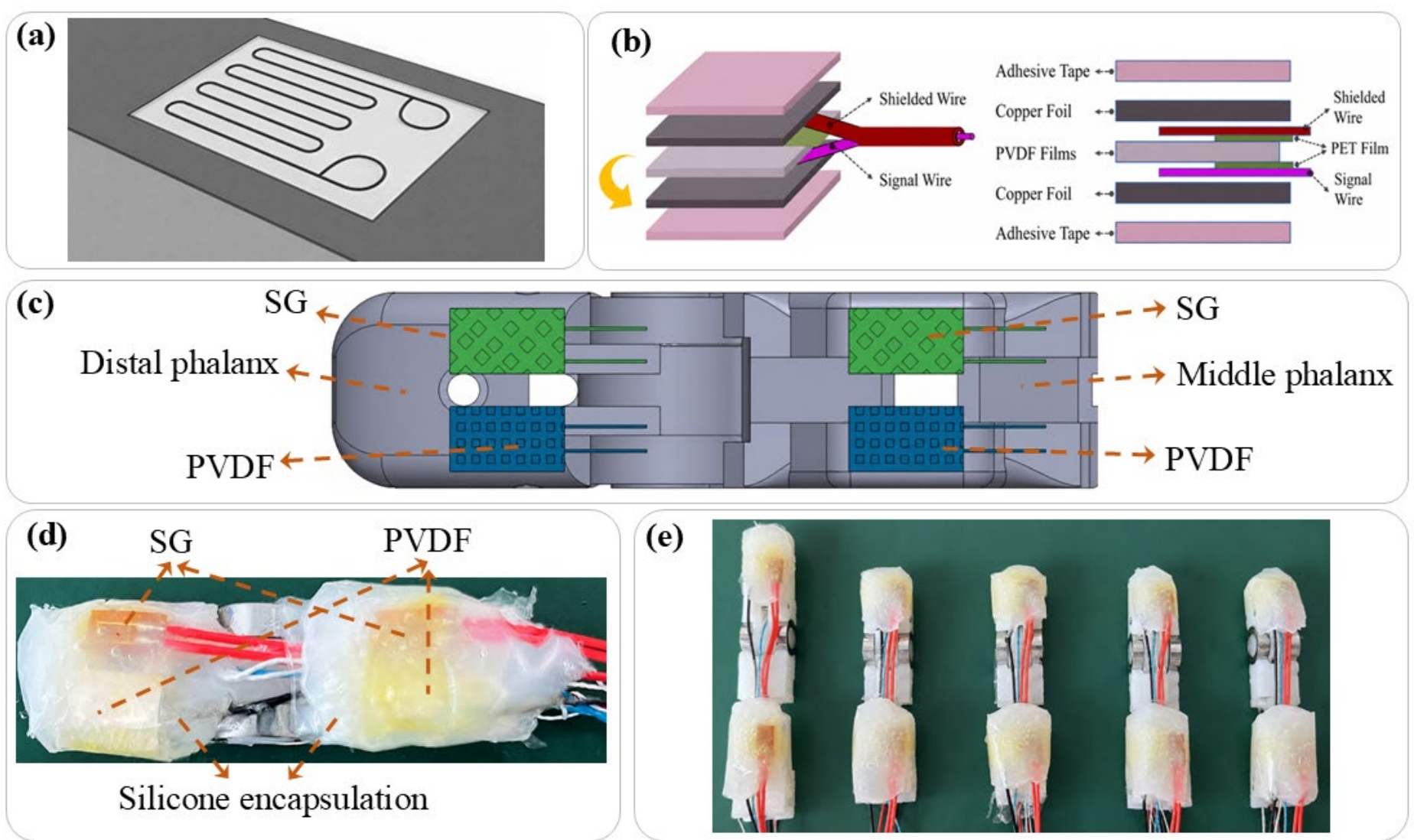

Fig. 3. Design of the distributed tactile perception on the dexterous hand. (a) Static tactile SE: strain gauge. (b) Dynamic tactile SE: PVDF. (c) Tactile SEs layout on each finger. (d) Fabricated soft-rigid-hybrid tactile finger. (e) All five fabricated fingers.

To simulate human perception underneath the skin, we embedded two types of tactile sensing elements (SEs) into the soft skin of the hand. Specifically, strain gauges (SGs) as shown in Fig. 3(a), a piezoresistive SE, were utilized to mimic the slow-adapting (SA) type of mechanoreceptors, which respond to static stimuli [9], [10]. Meanwhile, polyvinylidene fluorides (PVDFs) as shown in Fig. 3(b), a piezoelectric SE, were leveraged to play the similar role as fast-adapting (FA) type of mechanoreceptors, i.e., response to dynamic stimuli [11], [12]. In the soft skin of each finger, a pair of SG and PVDF SEs was attached to the distal phalanx while another pair to the middle phalanx. Thus, both segments are capable of perceiving both static and dynamic contact forces, and there are four tactile SEs in a finger, as illustrated in Fig. 3(c). The tactile finger after fabrication can be found in Fig. 3(d) while all five fingers are shown in Fig. 3(e).

### 3.2 Principles of the Tactile Perception

Different from existing tactile fingertips, all the five full fingers are endowed with the capability of dual-modality tactile perception. As for the piezoresistive SE, the deformation induced by any stimulus will lead to a change of its resistance. After conversion to voltage, this variance could be quickly captured. Considering SG primarily responds to relatively slow variance of applied forces, it is mainly utilized to sense the static or quasi-static signals which possess low frequency. Such applications include grasp and release, button press, object lifting, collision, and continuous contact etc. In contrast, PVDF is mainly responsible to dynamic stimuli. As a self-powered SE, different magnitude of voltages is generated between its both sides when various dynamic forces are applied. After filtering and amplification, the voltage is directly acquired by an ADC (analog-to-digital) module at a sampling rate of 1000 Hz. Its applicable scenarios comprise sliding, tap, slip detection and other fast-adapting requirements since it could respond to higher frequency components.

## 4. Experimental Evaluations

### 4.1 Manipulation Tasks for Dexterity Demonstration

Three manipulation tasks were designed to evaluate the dexterity of our developed robotic hand. In the first set of manipulations, the dexterous hand is controlled to make five different counting gestures as shown in Fig. 4(a). While some fingers are driven to implement the flexion movement, the others stay straight representing the number 1~5. It could be found that either the flexion or the extension movement could be implemented correctly, highly resembling the gestures of human hand.

As a typical operation, pinch action is critical in plenty of fine manipulations, such as picking up small screws or threading a needle. For human hands, the thumb is able to touch and implement pinch operations with any of the four remaining fingers. Similarly, four pinch operations were explored with our developed hand and the result was given in Fig. 4(b). In the synergy of finger driving tendon and swing tendon, high dexterity is endowed to the thumb and the four pinch operations are also implemented successfully.

The third set of manipulation tasks comprises three grasp operations: bottle grasp, squeeze action, and pinch and holding. In the bottle grasp manipulation, all five fingers executed the flexion movement after a plastic bottle was placed near the palm. The squeeze action is useful in gripping a thin card or cigarette. In this task, the index and middle finger is dominantly driven by the swing tendons while the other three fingers stay in bent status. In the pinch and holding task, the thumb and index finger are controlled to pinch a card and then keep the holding gesture while the other three fingers unbend during whole process. The successful implementation of all these tasks demonstrated that the designed hand was competent in making a large set of different gestures, which were vital in constituting various action primitives.

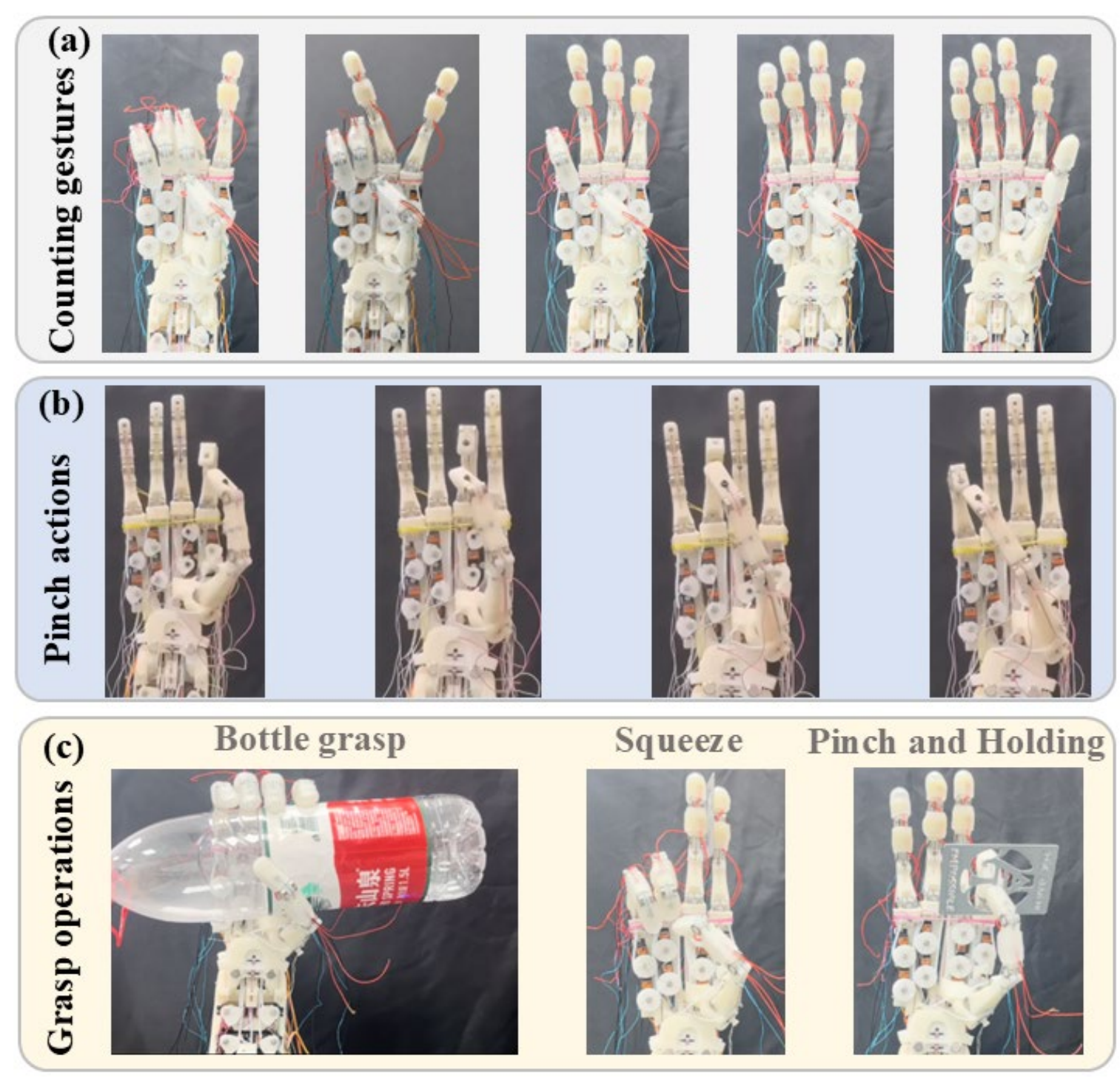

Fig. 4. Functional demonstrations of the designed dexterous hand with different manipulation tasks. (a) Five different counting gestures respectively representing the number of 1~5. (b) Four pinch actions between thumb and other four fingers. (c) Three typical grasp operations: bottle grasp with five fingers, squeeze action between index and middle finger, and pinch and holding operation using thumb and index finger.

### 4.2 Tactile Perception in Grasp Manipulations

To evaluate the tactile perception performance of our designed tactile module, we collected the tactile signals on all channels during a bottle grasp task. In Fig. 5, the four-channel tactile signals on thumb finger are displayed while the five primary phases, i.e., grasp, lifting up, holding, lifting down and release phase, are highlighted by different colored background.

It can be observed in general that 1) two PVDF channels primarily respond to the transition stage between two contact states, such as from non-contact to contact, from still to sudden movement, and from grasp to release. Two SG channels present relatively slow variation during whole manipulation process, and they show apparent responses in each phase when the applied static force changes. 2) There exists a time lag between the middle and distal segment both for PVDF and SG channels since the contact status between the middle segment and grasped bottle varies earlier. 3) Although the applied force direction cannot be determined, the grasping force that is normal to the finger surface and the lifting force that is tangential to the finger surface will be superposed in the magnitude, as indicated by the lifting up and down phases.

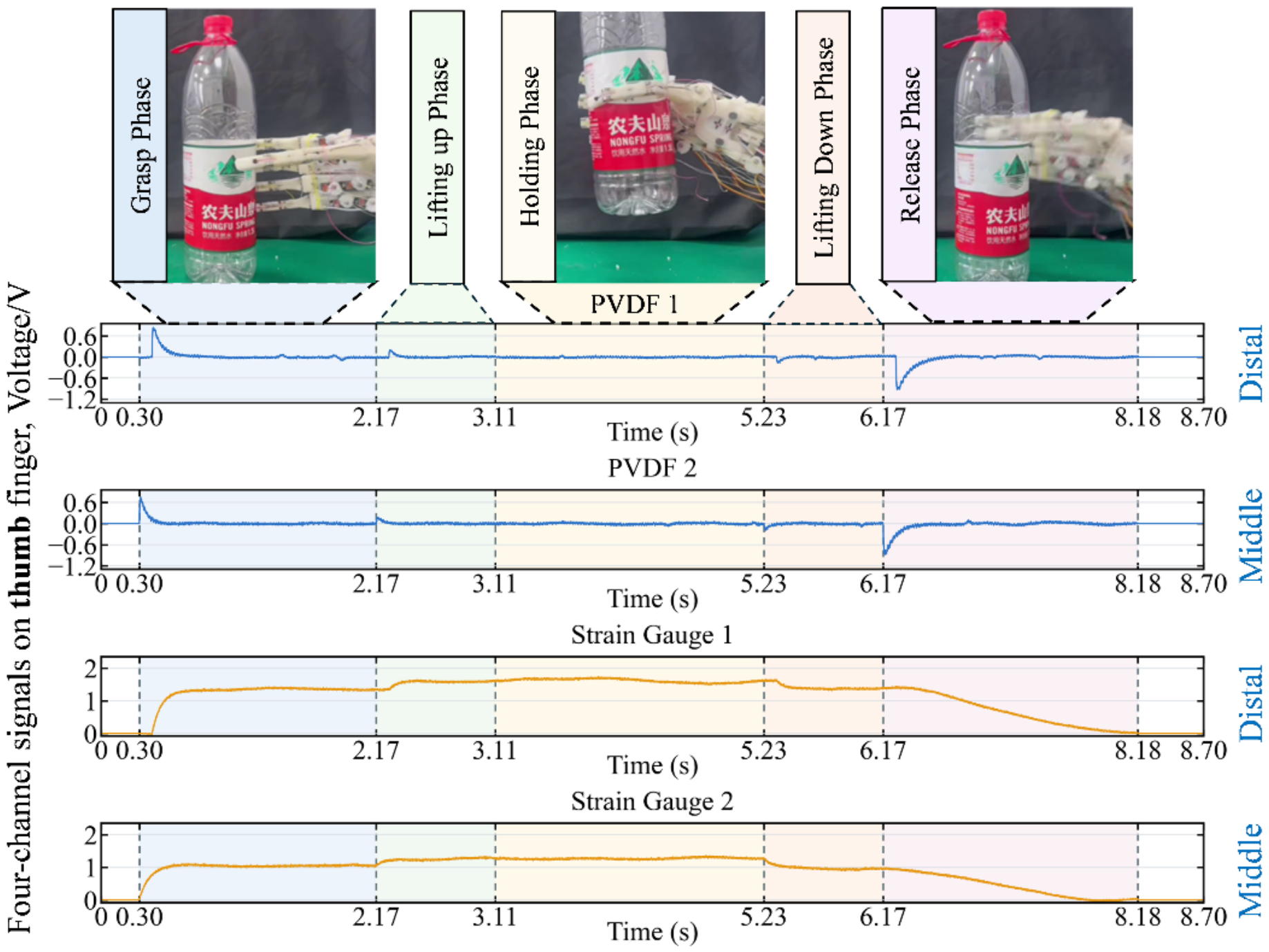

Fig. 5. Four-channel tactile signals on thumb finger in the manipulation task of bottle grasp, which mainly consists of five phases: grasp, lifting up, holding, lifting down and release. The four panel curves respectively show the PVDF1, PVDF2, SG1 and SG2 signals on distal and middle segment.

## 5. Conclusion

This paper presented a tendon-driven five-fingered hand capable of distributed dual-modality tactile perception. All involved primary components were detailedly described including the mechanical design, tactile sensor, driving mode and control system etc. Through a series of manipulation tasks, the dexterity and manipulation performance were successfully validated, showing its great potential in robotic manipulation. Finally, the tactile perception capability was revealed via in-depth analysis on the four-channel tactile signals in a bottle grasp task. Next, we plan to apply the dexterous hand to more complex applications, by developing more action primitives and taking full advantages of the tactile perception capability. Meanwhile, a variety of state-of-the-art techniques on embodied artificial intelligence can be combined to improve the operation performance and dexterity.

## References

[1] Y. Zhang and Z. Wang, "Review of robotic grippers for high-speed handling of fragile foods," *Adv. Robot.*, vol. 39, no. 17, pp. 1054–1070, Sep. 2025.

[2] Y. Shan *et al.*, “Variable stiffness soft robotic gripper: design, development, and prospects,” *Bioinspir. Biomim.*, vol. 19, no. 1, p. 011001, Jan. 2024.
[3] L. Qin, “Recent progress and challenges of key technologies in robotic assembly,” *Chin. J. Mech. Eng.*, vol. 39, p. 100032, Jan. 2026.
[4] J. Luo, C. Xu, J. Wu, and S. Levine, “Precise and dexterous robotic manipulation via human-in-the-loop reinforcement learning,” *Sci. Robot.*, vol. 10, no. 105, p. eads5033, Aug. 2025.
[5] O. M. Andrychowicz *et al.*, “Learning dexterous in-hand manipulation,” *The International Journal of Robotics Research*, vol. 39, no. 1, pp. 3–20, Jan. 2020.
[6] “DexHand mechanical build,” DexHand Project. [Online]. Available: https://www.dexhand.org/
[7] Y. Wang, T. Hao, Y. Liu, H. Xiao, S. Liu, and H. Zhu, “Anthropomorphic Soft Hand: Dexterity, Sensing, and Machine Learning,” *Actuators*, vol. 13, no. 3, p. 84, Feb. 2024.
[8] W. Yang, H. Zhao, C. He, and L. Qin, “Electrical Connector Assembly Based on Compliant Tactile Finger with Fingernail,” *Biomimetics*, vol. 10, no. 8, p. 512, Aug. 2025.
[9] L. Qin, X. Shi, Y. Wang, and Z. Zhou, “Perception of Static and Dynamic Forces with a Bio-inspired Tactile Fingertip,” *J Bionic Eng*, vol. 20, no. 4, pp. 1544–1554, Jul. 2023.
[10] L. Qin, X. Shi, W. Yang, Z. Qin, Z. Yi, and H. Shen, “Surface Recognition With a Tactile Finger Based on Automatic Features Transferred From Deep Learning,” *IEEE Trans. Instrum. Meas.*, vol. 73, pp. 1–10, 2024.
[11] X. Shi, Y. Wang, and L. Qin, “Surface Recognition With a Bioinspired Tactile Fingertip,” *IEEE Sensors J.*, vol. 23, no. 16, pp. 18842–18855, Aug. 2023.
[12] H. Zhao, X. Shi, W. Yang, H. Chen, and L. Qin, “Tactile Exploration-Enabled Shape Recognition With Multiperspective Feature Representation,” *IEEE Sensors J.*, vol. 25, no. 20, pp. 38780–38791, Oct. 2025.